\documentclass[]{article}

\usepackage[dvips,final]{graphicx} 

\usepackage{caption} 
\usepackage{subcaption} 
\usepackage{wrapfig} 
\usepackage{float}

\usepackage[dvipsnames]{xcolor} 
\usepackage{parskip} 
\usepackage[numbers,sort&compress]{natbib} 
\usepackage{titlesec} 
\usepackage{amsmath,amsfonts,amssymb,bm} 
\usepackage{times}
\usepackage{latexsym}
\usepackage{hhline}
\usepackage{soul} 
\usepackage{booktabs} 
\usepackage{colortbl} 

\usepackage[T1]{fontenc} 
\usepackage{array} 

\title{A spectrum of physics-informed Gaussian processes for regression in engineering} %
\author{E. J. Cross, T. J. Rogers, D. J. Pitchforth, S. J. Gibson and M. R. Jones$^1$.}

\date{%
$^1$	Dynamics Research Group, Department of Mechanical Engineering,\\
	University of Sheffield, Mappin Street, Sheffield S1 3JD, UK.\\ [2ex]%
}

\begin{document}
	
\definecolor{shadecolor}{rgb}{1,0.8,0.3}

\maketitle

\section*{Abstract}
Despite the growing availability of sensing and data in general, we remain unable to fully characterise many in-service engineering systems and structures from a purely data-driven approach. The vast data and resources available to capture human activity are unmatched in our engineered world, and, even in cases where data could be referred to as ``big,'' they will rarely hold information across operational windows or life spans. This paper pursues the combination of machine learning technology and physics-based reasoning to enhance our ability to make predictive models with limited data. By explicitly linking the physics-based view of stochastic processes with a data-based regression approach, a spectrum of possible Gaussian process models are introduced that enable the incorporation of different levels of expert knowledge of a system. Examples illustrate how these approaches can significantly reduce reliance on data collection whilst also increasing the interpretability of the model, another important consideration in this context.

\section*{Impact statement}
The availability of monitoring data from engineering structures offers many opportunities for optimising their design and operation. The ability to be able predict the current health state of a structure, for example, opens the door to predictive maintenance and, in turn, enhanced safety and reduced cost and waste. Currently most attempts at harnessing knowledge within collected datasets are reliant on those data alone, limiting the potential of what can be done to how much of the entire life of a structure is captured. This paper demonstrates how using physical knowledge within a machine learning approach can improve predictions considerably, reducing the burden on expensive data collection. A range of approaches allow differing levels/types of physical insight to be incorporated. 


\section{Data vs physics - an opinionated introduction} 
\label{sec:intro} 





The umbrella term \textit{data-centric engineering}, and interest in it, results from our growing ability and capacity to collect data from our built environment and engineered systems. The potential gains from being able to harness the information in these data are large, attracting many researchers and practitioners to the field. Here, we enjoy the very engaging challenge of simultaneously assessing what we think we know, what we can measure and what that might actually tell us about the particular structure or system that we are interested in. 

As our data grow, many researchers naturally look to adopt machine learning (ML) methods to help analyse and predict behaviours of interest in our measured systems. Within this vibrant field, there are many advances with the potential to enhance or optimise how we design and operate our human-made world. The drivers and challenges of taking a data-driven approach in an engineering setting are often, however, different for those developing the latest ML/AI algorithms. Generalising significantly, where a machine learning practitioner might look to develop a powerful algorithm that can make predictions across many applications without user input, an engineer's interest may necessarily be more system-focussed and should also consider what knowledge can be gained from the model itself alongside optimising its predictive capability. In terms of challenges, one of the most significant is simply that engineering data, although sometimes ``big'', often do not capture all behaviours of interest, may consist of indirect measurements of those behaviours and, for operational monitoring, will often be noisy, corrupted or missing. 

To give an example where these challenges are particularly pertinent, consider the problem of asset management of civil infrastructure. Given a fixed budget for a monitoring system, we'd like to be able to collect and use data to tell us something about the current condition of a structure (or multiple structures), ultimately we'd really like to be able to use that data to make predictions about how the structure(s) will perform in the future. As a monitoring system comes online, the data available may be large in size, however, the information content in it will be limited by the operational conditions seen in the monitoring window, the current condition of the structure and the robustness of the sensing and acquisition system. A data-driven model established in this setting bears the same limitations and we should be careful (and are) about how we expect any such model to generalise to future structural and operational conditions. 

Many have and will question why pursuing a data-driven approach is interesting in this context given these limitations/challenges, especially considering our efforts and successes throughout history in understanding and describing the world and beyond via physics. The blunt answer is that we don't know everything and, when we do know enough, modelling complex (multi-physics, multi-scale) processes interacting with a changing environment is often difficult and energy consuming. The more compelling answer, perhaps, is that we would really like any inferences we make to be based on evidence of what is happening currently, and for that, observation is needed.

\begin{figure}
	\centering
	\includegraphics[width=0.5\linewidth]{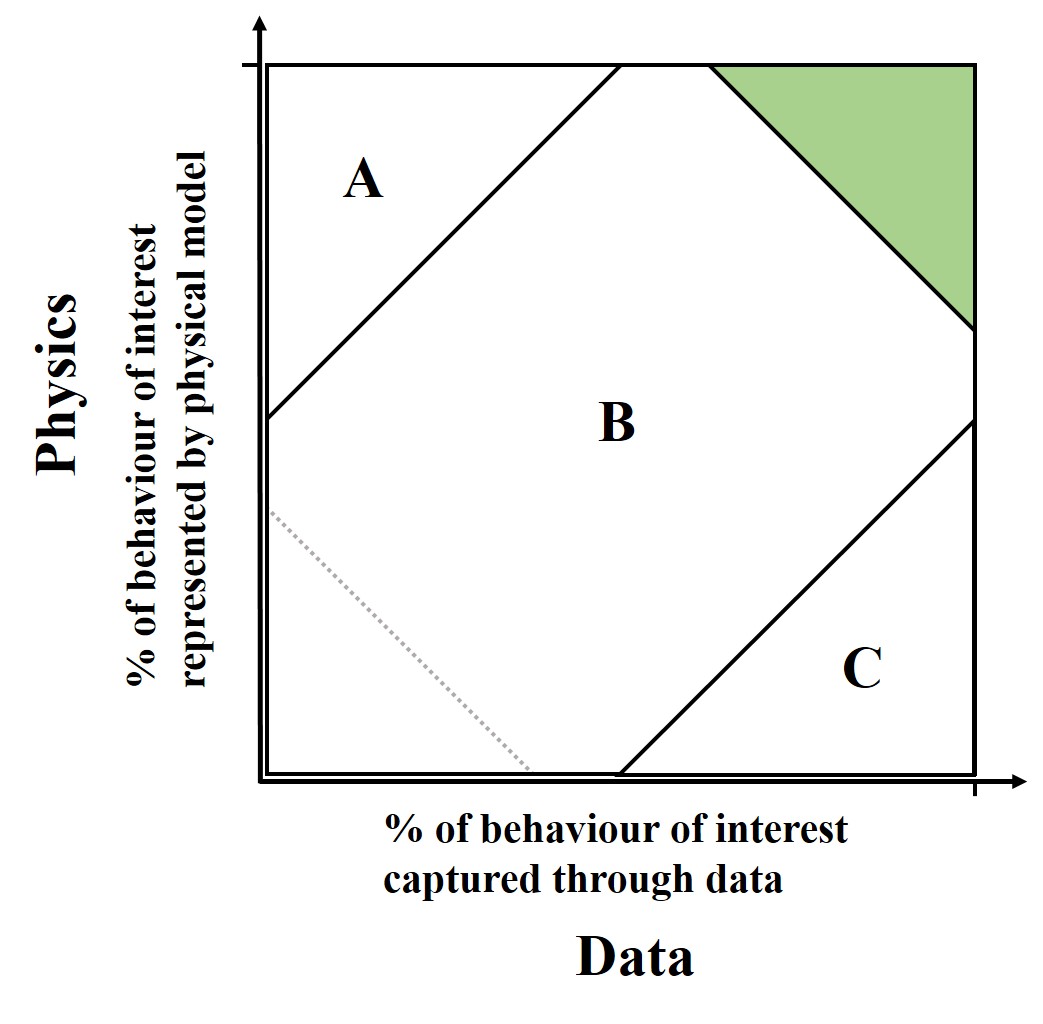}
	\caption{A mapping of problem settings according to knowledge available from physical insight and data}
	\label{fig:spec}
\end{figure}

Given a particular context, there are, of course, benefits and disadvantages in taking either approach; a physics-based one can provide interpretability, parsimony and the ability to extrapolate, for example, whereas data-driven approaches can be considered more flexible, may require less user input, and take into account evidence from latest measurements. The selected route may come down to personal choice but hopefully is the result of a reasoned argument.

One hypothetical approach to this reasoning may be to consider how much of the process of interest can be described by known/modelled physics (within a given computational budget) versus how much of it can by characterised by available data (supposing one could assess such things) - see Figure \ref{fig:spec}. Broadly speaking, in region A one would likely take a physics-based approach and in region C a data-driven one. In the happy green area either may be applicable, and in the bottom left hand corner, one may wish to consider the best route to gain additional knowledge, whether through measurement or otherwise (the triangle sizes here are arbitrary and for illustration only).

Recently interest has been growing in methods that attempt to exploit physics-based models and evidence from data together, hopefully retaining the helpful attributes from both approaches. Indeed, there has been an explosion of literature on these methods in the last few years, duly reviewed in \cite{vonrueden23informed,karniadakis2021physics,willard2020integrating}. These methods should be useful for the many engineering problems that fall in region B in Figure \ref{fig:spec} - the grey area. `Grey' is also occasionally used to describe the models themselves, alluding to them being a mix of physics (so called `white-box') and data-driven (so called `black-box') approaches. The term `physics-informed machine learning' or similar is now also commonly used for those methods at the darker end of the scale.

\begin{figure}
	\centering
	\includegraphics[width=0.9\linewidth]{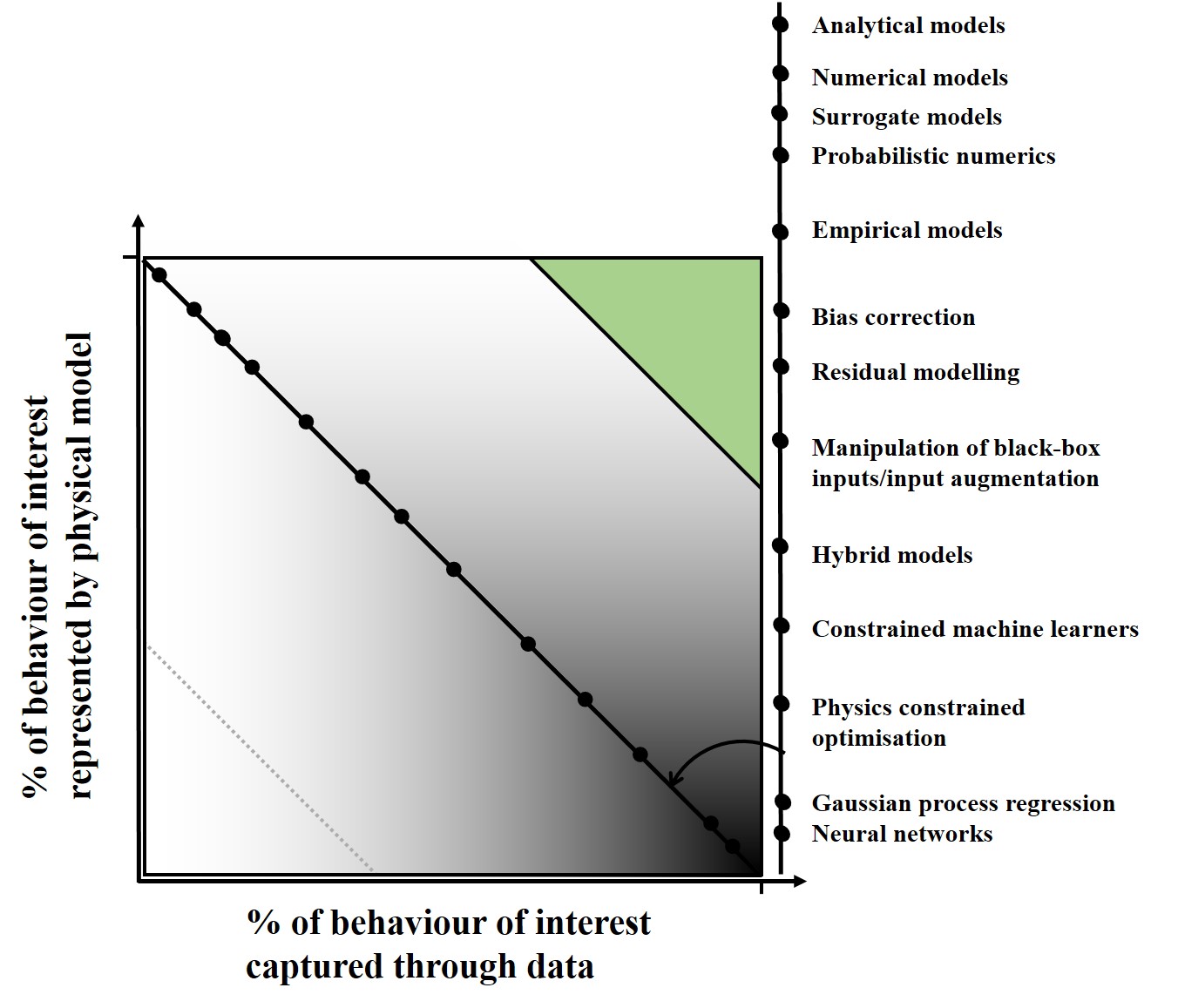}
	\caption{A non-exhaustive list of modelling approaches very loosely mapped onto the data/physics problem setting axes from Figure \ref{fig:spec}.}
	\label{fig:spec_lit}
\end{figure}

Figure \ref{fig:spec_lit} very loosely maps some of the available methods onto the knowledge from physics and data axes from the original figure and the black to white spectrum (of course it must be noted here that method location will change according to the specific model type and how it is applied - sometimes significantly so). At the lighter end of the scale, surrogate models are a continually growing area of interest \cite{kennedy2001bayesian,QUEIPO20051,bhosekar2018advances,ozan2022physics} where emulation of an expensive to run physical model is needed. The emerging field of probabilist numerics \cite{cockayne2019bayesian,hennig2022probabilistic} attempts to account for uncertainty within numerical modelling schemes and overlaps with communities specifically looking at bias correction and residual modelling \cite{gardner2021learning,brynjarsdottir2014learning,arendt2012quantification}, where a data-driven component is used to account for error in a physical model in the first case, or behaviours not captured by a physical model in the latter (most often achieved by summing contributions from both elements). In the middle sit methods where there is a more even share between the explanatory power of physics and data-driven components; this may simply be achieved by feeding the predictions of a physical model into a machine learner as inputs \cite{rogers2017grey,worden2018evolutionary,fuentes2014aircraft}, or may involve a more complicated architecture informed by understanding of the process itself. Examples of these hybrid models include neural networks where the interaction between neurons and the design of their activation functions reflect knowledge of the physics at work in the situation to be modelled \cite{cai2021physics, lai2021structural}. Finally at the blacker end of the scale sit constrained methods where, for example, laws or limits can be built into a machine learner to aid optimisation or ensure physically feasible predictions. Many studies in this area focus on constraining a cost function \cite{cai2021physics,Karpatne2017} for e.g.\ parameter optimisation, with fewer focussing on directly constraining the model itself (as will be the case in some of this work) \cite{solin2018modeling, wahlstrom2013modeling,jidling2018probabilistic,jones2023constraining}.



This paper introduces a spectrum of models from white to black under a Gaussian process (GP) prior assumption. Gaussian processes have been shown to be a powerful tool for regression tasks \cite{Rasmussen2006}, and their use in this context within engineering is becoming common (see e.g. \cite{kullaa2011distinguishing,avendano2017gaussian,wan2018bayesian,wan2019bayesian}). The regular use of GP regression by the authors of this paper (e.g. \cite{cross2012structural,holmes2016prediction,rogers2020application,bull2020probabilistic}) is because of their flexible yet simple nature, ability to function given small datasets and, importantly, the Bayesian framework within which they naturally work; the predictive distribution provided allows the calculation of useful confidence intervals and the opportunity for uncertainty to be propagated forward into any following analysis (see \cite{gibson2020data}, for example). Despite these advantages, their use in the provided citations remains entirely data-driven and thus open to the challenges/limitations discussed above.

By drawing an explicit link between the classical treatment of GPs in physical sciences and how they are used for regression in a machine learning context, this paper illustrates a number of potential routes for combining differing depths of physical reasoning with learning from data. To do so, Section \ref{sec:SP} introduces Gaussian processes from both perspectives, with Section \ref{sec:spec} identifying their overlap and the specific means of incorporating physical insight into a GP regression, which are then illustrated and further discussed through examples across the spectrum in Section \ref{sec:examples}. Finally Section \ref{sec:discussion} discusses the proposed models in reference to other relevant examples in the literature and draws conclusions on their future use.

\section{GPs from a physics and machine learning perspective}
\label{sec:SP}
{\bf Physics-based perspective} \\ 
From the classical perspective, a Gaussian process is one example in a wider family of stochastic processes used to characterise randomness. To describe a stochastic process intuitively, to begin, one can consider 
something that evolves through \textit{time}. In this case, a stochastic 
process is one where, at each instance of time, $t$, the value of the process 
is a random variable. In characterising the stochastic process, we are 
describing the evolution of probability distributions through time. Note that 
we may equally wish to consider the 
evolution of a process through a variable ($\in \mathbb{R}^n$) that is not 
time and will do so later in the paper.




The fundamental elements for describing a stochastic process are the mean and 
autocorrelation, which are functions over time (or the input variable(s) of 
interest). Considering a process ${y}(t)$, its mean $\mu(t)$ and 
autocorrelation $\phi(t_1,t_2)$ functions are
\begin{equation}
\begin{aligned} 
\mu(t)=&\mathbb{E}[{y}(t)] \\
\phi(t_1,t_2)=&\mathbb{E}[{y}(t_1){y}(t_2)] \\
=\int_{-\infty}^{\infty}\int_{-\infty}^{\infty} 
{y}(t_1)&{y}(t_2)g({y}(t_1),{y}(t_2)) 
dy(t_1)dy(t_2)
\end{aligned}
\label{eq:sp_mean_autocor}
\end{equation}
where $\mathbb{E}$ is the expectation operator. The autocorrelation requires 
integration of the product of ${y}(t_1){y}(t_2)$ and their joint probability 
density, $g$, at times $t_1$ and $t_2$ (this is sometimes referred to as the 
second order density).

Following on from this, the (auto)covariance of a process, $k(t_1,t_2)$, is
\begin{equation}
k(t_1,t_2)=\mathbb{E}[({y}(t_1)-\mu(t_1))({y}(t_2)-\mu(t_2))]
\label{eq:cov}
\end{equation}
Clearly, the autocorrelation and (auto)covariance are one and the same for a process with a zero mean.

A \textbf{{\em  Gaussian} process} is one where at each instance or iteration, the value of the variable of interest follows a normal/Gaussian distribution, with the joint distribution of a finite collection of these also normal. It is completely defined by its mean and the covariance function, i.e. one need only consider the joint density between two points (second order density). 

Many of the behaviours/variables that we wish to model in science and engineering are stochastic processes. The first use of the term `stochastic process' arose in the 1930s \cite{khintchine1934korrelationstheorie,doob1934stochastic}, but the response of a physical system to random excitation, which is most certainly a stochastic process, had been under study since at least the turn of the 20th century.\footnote{In 1905 Einstein derived the probability distributions of the displacement through time of particles suspended in fluid \cite{einstein1905motion}. For the interested reader, two review papers on Brownian motion by Uhlenbeck and co-authors provide an excellent discussion of the work around this time \cite{uhlenbeck1930theory,wang1945theory}.}

A particular example of interest that will be used later in this paper is that of a structure vibrating. Perhaps the simplest 
formulation of a stochastic process in this context is the response of a 
deterministic system under random excitation. Generally 
speaking, if the excitation to a linear dynamical system is a Gaussian 
process, then the response of that system is also a Gaussian process (a 
Gaussian process remains a Gaussian process under linear operations  
\cite{papoulis2002probability}).  

Given an assumed equation of motion (or physical law of interest) for $y$, one can 
attempt to derive the covariance function of a process using 
(\ref{eq:sp_mean_autocor}, \ref{eq:cov}). Later in the paper the derived covariance function for a single degree of freedom (SDOF) system
under random loading will be shown and employed in a Gaussian process regression setting. 

{\bf Data-driven perspective} \\
In the context of Gaussian process \textit{regression}, i.e. from a data-driven perspective, the process is unknown (to be learned from data), and so the definition of the mean and covariance functions become a modelling choice. These choices form the \textit{prior} mean and covariance, which will be updated to the \textit{posterior} mean and covariance given observations of the process of interest.

\begin{figure}
	\centering
	\begin{subfigure}{.5\textwidth}
		\centering
		\includegraphics[width=\linewidth]{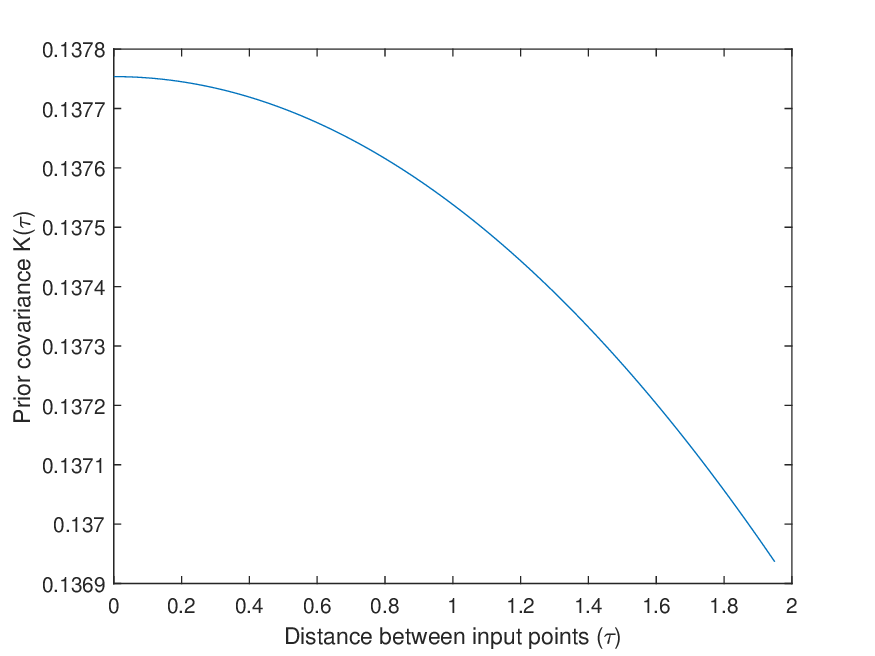}
		\caption{Squared-exponential}
		\label{fig:SE_prior_cov}
	\end{subfigure}%
	\begin{subfigure}{.5\textwidth}
		\centering
		\includegraphics[width=\linewidth]{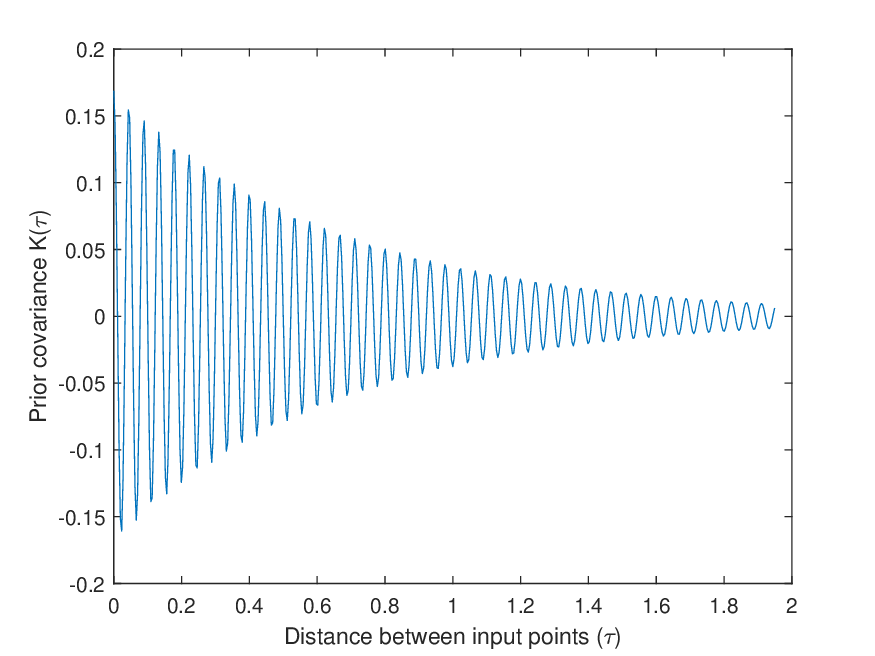}
		\caption{SDOF}
		\label{fig:SDOF_prior_cov}
	\end{subfigure}
	\caption{Measure of influence of an input point on a prediction for the 
		squared exponential (a) and the covariance function of single degree of 
		freedom (SDOF) oscillator under a random load (b).}
	\label{fig:prior_covs}
\end{figure}

An example of a common choice for covariance function is the squared-exponential with an additional white noise covariance term:
\begin{equation}
k({\bf x}_p,{\bf x}_q )=\sigma_y^2 \exp(-\frac{1}{2l^2} ||{\bf x}_p-{\bf x}_q ||^2 ) +\sigma_n^2\delta_{pq}	\label{eq:SE}
\end{equation}
Rather than a process through time, $t$, this is a covariance function for a process defined over a multivariate input space with elements ${\bf x}_i$ (reflecting the more generic nature of regression tasks attempted). Here there are three hyperparameters; $\sigma_y^2$, the signal variance, $l$, the length scale of the process and $\sigma_n^2$, the variance from the noise on the measurements. This may be easily adapted to allow a separate length scale for each input parameter if required.

After selection of appropriate mean and covariance functions and with access to measurements of the process of interest, the regression task is achieved by calculating the conditional distribution of the process at testing points given the observations/ measurements.

Following the notation used in \cite{Rasmussen2006}; $k({\bf{x}}_p,{\bf{x}}_q)$ defines a covariance matrix $K_{pq}$, with elements evaluated at the points ${\bf x}_p$ and ${\bf x}_q$, where ${\bf x}_i$ may be multivariate. 

Assuming a zero-mean function, the joint Gaussian distribution between measurements/observations $\bf{y}$ with inputs $X$ and unknown/testing targets ${\bf y}^*$ with inputs $X^*$ is 
\begin{equation}
\begin{bmatrix}
{\bf y} \\
{\bf y}^*
\end{bmatrix}
\sim \mathcal{N} \left( 0,
\begin{bmatrix}
K(X,X)+\sigma_n^2I &  K(X,X^*) \\
K(X^*,X) & K(X^*,X^*) 
\end{bmatrix}
\right)
\label{eq:Sjoint}
\end{equation}
The distribution of the testing targets ${\bf y}^*$ conditioned on the training data (which is what we use for prediction) is also Gaussian:
\begin{equation}
\begin{aligned}
{\bf y}^*|X_*,X,{\bf y} \sim \mathcal{N}&(K(X^*,X)(K(X,X)+\sigma_n^2I)^{-1} {\bf y}, \\
&K(X^*,X^* )-K(X^*,X)(K(X,X)+\sigma_n^2I)^{-1} K(X,X^*))
\end{aligned}
\label{eq:condcvfn1}
\end{equation}
See \cite{Rasmussen2006} for the derivation. The mean and covariance here are that of the posterior Gaussian process.

From (\ref{eq:condcvfn1}) one can see that the GP mean prediction at point 
${\bf x}^*$ is simply a weighted sum - determined by the covariance function - 
of the training points ${\bf y}$. Figure \ref{fig:prior_covs}a illustrates 
how 
the influence of a training point on a prediction decays as the distance in 
the input space increases when using the squared exponential covariance 
function (hyperparameters arbitrarily selected). This shows how the 
covariance between points with similar inputs will be high, as is entirely 
appropriate for a data-based learner. In the absence of training data in an area of the input space, the mean value 
of the GP will return to the prior mean (usually zero). An equivalent plot for the covariance function of an SDOF oscillator employed later (Section \ref{sec:examples}) is included for comparison, where one can see the oscillatory nature captured. 



\section{A spectrum of Gaussian processes for regression} 
\label{sec:spec}


Commonly, engineering applications of GP regression will follow a typical machine learning approach and adopt a zero mean prior and a generic covariance function selected from either the squared-exponential or Mat\'{e}rn kernel class \cite{Rasmussen2006}. In an upcoming summary figure, this will be denoted as $f(x)\sim\mathcal{GP}(0,k_{ML})$, with the subscript $ML$ denoting the machine learning approach as above. Although successful in many settings, these applications suffer from those same challenges discussed in the introduction - namely that the regression bears the same limitations as the dataset available with which to characterise the system/structure of interest.

Here, the incorporation of one's physical insight of a system into a GP regression is introduced as a means of lessening reliance on complete data capture. The GP framework provides a number of opportunities for accounting for physical insight, the biggest coming from definition of the prior mean and covariance functions. Following on from the previous section, perhaps the most obvious approach is to use physically-derived mean and covariance functions where available (eqns.(\ref{eq:sp_mean_autocor}, \ref{eq:cov}) above). If these can be derived, they may readily be used in the regression context, denoted as $f(x)\sim\mathcal{GP}(\mu_P,k_{P})$ ($P$ for physics) in the following.

\begin{figure}
	\centering
	\includegraphics[width=0.9\linewidth]{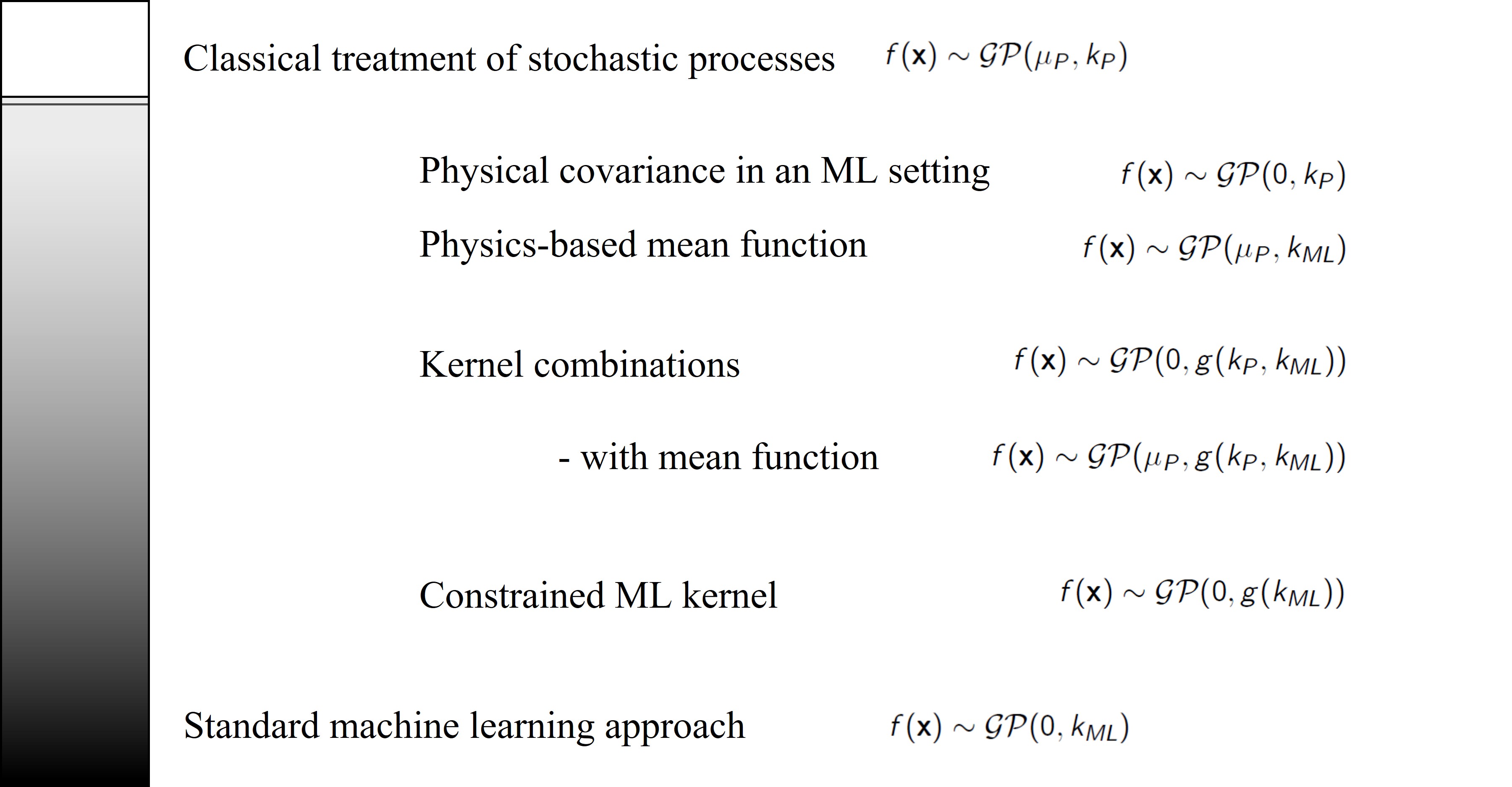}
	\caption{A spectrum of GPs for regression combining physics-derived mean and covariance functions, $\mu_P,k_{P}$, with those more standardly used in machine learning, $k_{ML}$.}
	\label{fig:spec_gp}
\end{figure}

This section lays out a number of modelling options for the more likely scenario that one has partial knowledge of the system of interest. Following the flow from more physical insight to less, each of the models discussed is placed on the white to black spectrum in Figure \ref{fig:spec_gp}, perhaps giving an indication of the kind of problem where they may be most usefully employed (illustrative examples of each follow in the next section). As with Figure \ref{fig:spec_lit}, it should be noted that how each model type is developed and applied could change its placement on the spectrum.




\textbf{Light grey} If one is able to express the mean behaviour of the process of interest, or something close to the mean, then this is easily accounted for by employing that mean as a prior with a standard machine learning covariance function to capture the variability around it: $f(x)\sim\mathcal{GP}(\mu_P,k_{ML})$ \cite{zhang2020gaussian,pitchforth2021grey}. If the form of the differential equation governing the process of interest is known and the covariance derivable, then this may also simply be used in place of the data-driven kernels discussed above, $f(x)\sim\mathcal{GP}(0,k_{P})$ \cite{cross2021physics}. In this case, the learning of unknown system parameters may be achieved by maximising the marginal likelihood, $p(y|X)$, in the way that one learns the hyperparameters in the standard machine learning approach (see for example \cite{Rasmussen2006}).

\textbf{Medium grey} In the more likely scenario of only possessing partial knowledge of the governing equations of a system of interest, or not being able to derive a covariance analytically, the suggestion here is that the GP prior may be approximated or formed as a combination of derived and data-driven covariance functions so that the data-driven component accounts in some way for unknown behaviour; $f({\bf x})\sim \mathcal{GP} (0,g(k_{P},k_{ML}))$, for some function $g$, or perhaps $f({\bf x})\sim \mathcal{GP} (\mu_P,g(k_{P},k_{ML}))$ if a mean may be appropriately approximated. 


Although this is an area very much still under investigation, the suggested route here is to begin, if possible, by considering or assuming the likely interaction between the known and unknown behaviours and to propagate this through to the prior GP structure. For example, say that one can assume that the response of a structure is a sum of understood behaviour and an unknown contribution; $y=A+B$, with $A$ known and $B$ unknown, then the autocorrelation of this process can be formed as

\begin{equation}
	\mathbb{E}[yy']=\mathbb{E}[(A+B)(A'+B')]=\mathbb{E}[AA'+AB'+A'B+BB']
\end{equation}

If $A$ and $B$ may be assumed independent and we make the standard machine learning prior assumption on $B$ of a zero mean and covariance $K_B$, this suggests a suitable GP model\footnote{A note of caution here: in considering a derivation path rooted in physical insight, we must be careful not to be fooled into thinking that the GP defined by mean and covariance derived from partial knowledge is representative of the underlying generative stochastic process, as in many cases this may not be Gaussian. The implicit assumption taken in any GP regression is that the target of interest can be suitably modelled by \textit{some} Gaussian process \textit{prior}, with the flexibility of the commonly adopted ML kernels allowing useful models under this assumption. The assumption, in turn, made here, is that the \textit{combination} of these kernels with ones derived from physical knowledge also describes some Gaussian process that can be used to model the target.} would be $y\sim\mathcal{GP}(0,K_A+K_B)$. As any linear operation between covariance functions is valid, this route is available for many likely scenarios in an engineering setting, for example, where the response to be modelled is a convolution between a known system and unknown force (as will be the case for the derived covariance in the first example below), or, e.g., a product between a known temporal response and unknown spatial one ($y=A(t)B(x)\implies$$y\sim\mathcal{N}(0,K_A(t)\times K_B(x))$, for $x$ and $t$ independent). 

\textbf{Dark grey} Finally, if the physical insight one has is perhaps more general or cannot be expressed through a mean or covariance function, one can consider adapting a data-driven covariance function to obey known constraints or laws. This may be done by, for example, the construction of a multiple-output GP with auto and cross-covariance terms designed to reflect our knowledge \cite{wahlstrom2013modeling,jidling2018probabilistic}, or by constraining predictions onto a target domain such that boundary conditions on a spatial map can be enforced \cite{solin2019know, jones2023constraining}. 

The next section shows examples for each of these categories, with discussion following in Section \ref{sec:discussion}. Each of the examples is presented quite briefly, with references for further reading. The intention, and the reason for brevity, is to attempt to illustrate models across the range of spectrum with conclusions drawn from them jointly in Section \ref{sec:discussion}.

\section{Examples}
\label{sec:examples}

\subsection{Physics-derived means and covariances}
The light grey models described above are applicable when one can derive an approximate mean or covariance function for the process of interest. 

{\bf Example 1} $f({\bf x})\sim \mathcal{GP} (0,k_{P})$ 

In Section \ref{sec:SP}, an oscillatory system under white noise was used as an example of a Gaussian process in time. For a single degree of freedom (SDOF), the equation of motion is $	m\ddot{{y}}(t)+c\dot{{y}}(t)+k{y}(t)={F}(t)$, with $m$, $c$, $k$ and $F$ the mass, damping, stiffness and force respectively. Under a white noise excitation of variance $\sigma^2$, one can derive the (auto)covariance of the response $Y(t)$; $\phi_{Y(t_1)Y(t_2)}=\mathbb{E}[Y(t_1)Y(t_2)]$, which is solvable either via some lengthy integration or by Fourier transform of the power spectral density \cite{cross2021physics}:


\begin{equation}
	\phi_{Y}({\bf 
		\tau})=\frac{\sigma^2}{4m^2{\zeta\omega_n}^3}e^{-\zeta\omega_n|\tau|}(\cos(\omega_d\tau)+\frac{\zeta\omega_n}{\omega_d}\sin(\omega_d|\tau|))
	\label{eq:mean_autocorr_sdof}
\end{equation}

Here standard notation has been used; $\omega_n=\sqrt{k/m}$, the natural frequency, $\zeta=c/2\sqrt{km}$, the damping ratio, $\omega_d=\omega_n\sqrt{1-\zeta^2}$, the damped natural frequency and $\tau=t_i-t_j$ (this is a stationary process). 

\begin{figure}
	\centering
	\includegraphics[width=0.8\linewidth]{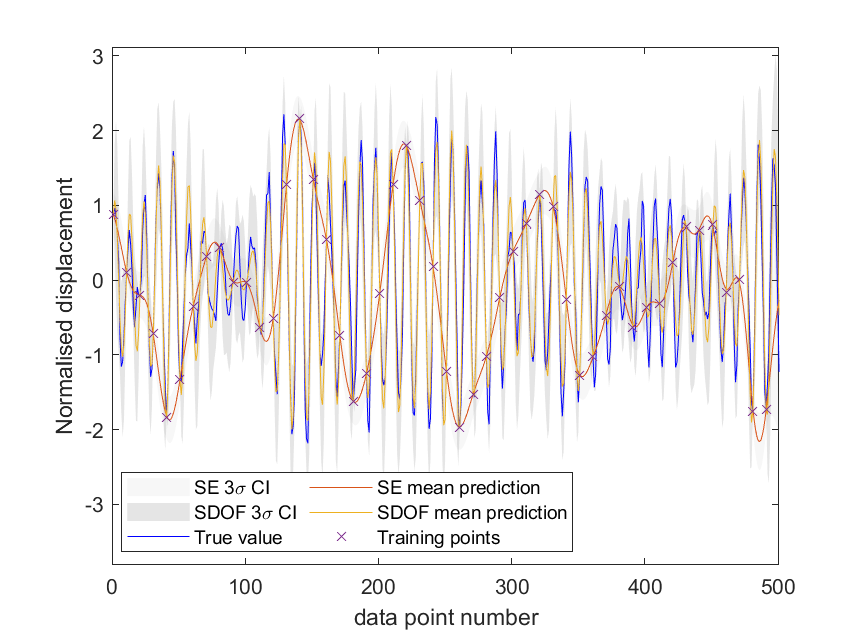}
	\caption{Comparison between the prediction of GPs with squared-exponential (SE) and SDOF kernels when conditioned on every 10th point of simulated vibration data \cite{cross2021physics}. The grey area indicates confidence intervals (CI) at three standard deviations. }
	\label{fig:compare_SDOF_SE}
\end{figure}

If one has a system that behaves similarly to this, then we may readily use such a covariance function in a regression context. Figure \ref{fig:compare_SDOF_SE} shows the prediction of an undersampled SDOF system using a GP regression with this covariance function, compared to one with a more standard kernel in a machine learning setting (the squared-exponential). The crosses mark the training/conditioning points.

Hyperparameters for both models were learned from maximising the marginal likelihood, $p(y|X)$, with the hyperparameters of the SDOF kernel being the parameters, $\omega_n,\zeta$, of the system itself.

One can see that the GP with the standard covariance function (labelled as SE) smooths through the observed data as designed, and that the GP with the derived covariance function (labelled as SDOF mean prediction) is much more appropriately equipped to model the process than its purely data-driven counterpart. In particular, the inbuilt frequency content of the derived covariance function gives a significant advantage ($w_n$ is learned as a hyperparameter within a bounded a search space). 

In giving the kernel structure pertinent to the process of interest, one is able to significantly reduce reliance on conditioning data, in this case allowing sampling below Nyquist. This covariance function will also be employed in a later example with extension to multiple degrees of freedom. 

{\bf Example 2} $f(x)\sim\mathcal{GP}(\mu_P,k_{ML})$

In some situations, the mean behaviour of a process may be broadly understood. This example shows the prediction of deck displacement of a stay-cabled bridge\footnote{The bridge in question is the Tamar Bridge in Southwest England and actually has both stay and suspension cables.}, the model of which is intended for use in performance monitoring the structure. The deck displacement is a function of a number of drivers, principally traffic loading  and temperature.  Considering physical insight, we believe that the general displacement trend is driven by the contraction and relaxation of the cables with temperature, with seasonal trends visible. Here, therefore, a good candidate prior mean function is a linear relationship between cable extension and temperature. 

\begin{figure}
	\centering
	\includegraphics[width=\linewidth]{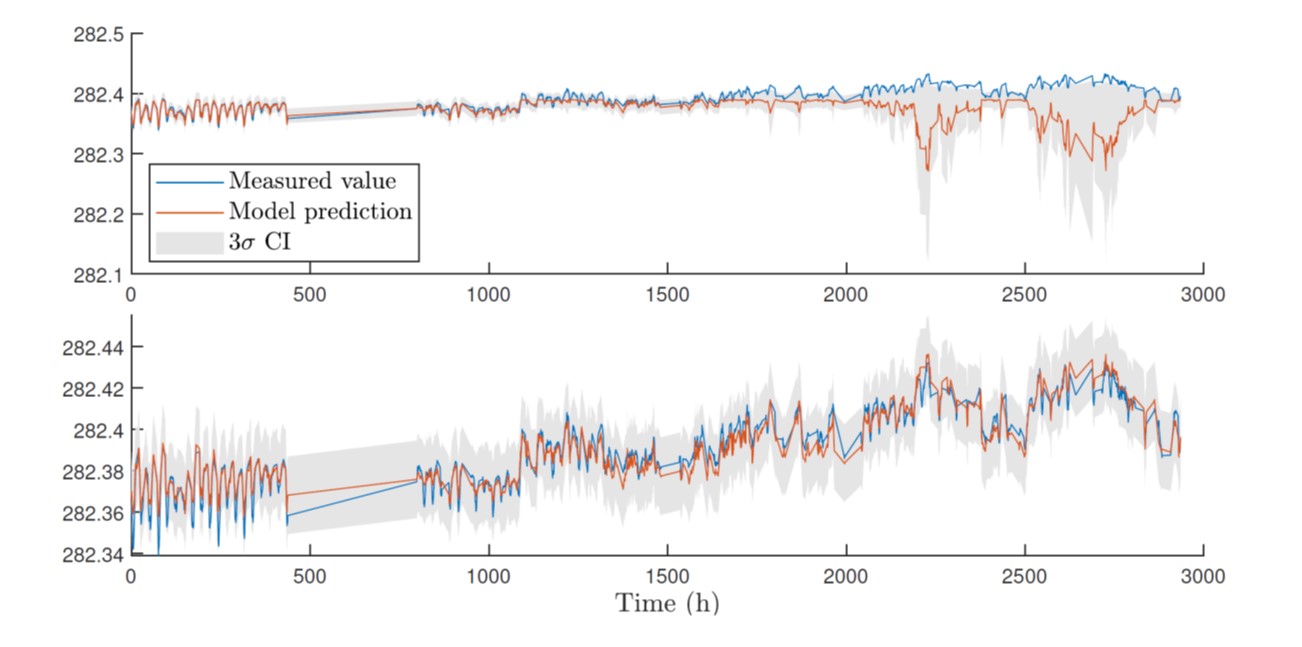}
	\caption{Comparison between GPs predicting bridge deck displacement over time with prior mean of zero above and with physics-informed mean function below \cite{zhang2020gaussian}.}
	\label{fig:compare_tamar}
\end{figure}

Figure \ref{fig:compare_tamar} compares two standard GP regressions (with squared-exponential covariance functions) with and without the prior mean function included. Mimicking the case where data from a full monitoring campaign is only available over a short time window, the models are trained (conditioning and hyperparameter setting) using data from the first month of the five month period shown. The GP with a zero mean prior (top image in Figure \ref{fig:compare_tamar}) is unable to accurately predict the deck displacement mean-wise as the temperature drops seasonally toward the end of the five month period (as this is an unseen condition, as indicated by the increased confidence interval). 

In this case, building in the linear relationship between cable extension and temperature as a prior mean function allows a more successful extrapolation into the colder months (lower image in Figure \ref{fig:compare_tamar}), again demonstrating a lessened reliance on complete data for training.

\subsection{Combined derived and data-driven covariance functions}
In the middle of the spectrum are the many problems where governing equations are only available to describe some of the behaviours of interest. The hybrid models here are proposed to account for this, with the interaction between physical and data-driven components necessarily more interwoven.  

{\bf Example 3} $f({\bf x})\sim \mathcal{GP} (0,g(k_{P},k_{ML}))$

The motivating problem here is the health/usage monitoring of an aircraft wing during flight, where we would like to predict wing displacement spatially and temporally to feed into a downstream fatigue damage calculation \cite{gibson2023distributions,holmes2016prediction}. In this case, one could assume that the solution of the equation of motion has separable spatial and temporal components, as with a cantilever beam. Under a random load, the covariance of the temporal component may be derived as in Example 1 (assuming linearity), accounting for multiple degrees of freedom by adding covariance terms up for each of the dominant modes (see \cite{pitchforth22incorporation}). As the wing will likely be complex in structure, a data-driven model component (covariance) is a good candidate to account for spatial variation.\footnote{If the wing was idealised as a cantilever beam, the covariance of the spatial component may also be readily derived.} With separability,
the covariance of the process will be a product between the spatial and temporal components, which would make the assumed GP model:  $y\sim\mathcal{GP}(0,K_{MDOF}(t)K_{ML}(x))$.

\begin{figure}
	\centering
	\begin{subfigure}{0.45\linewidth}
		\includegraphics[width=\linewidth]{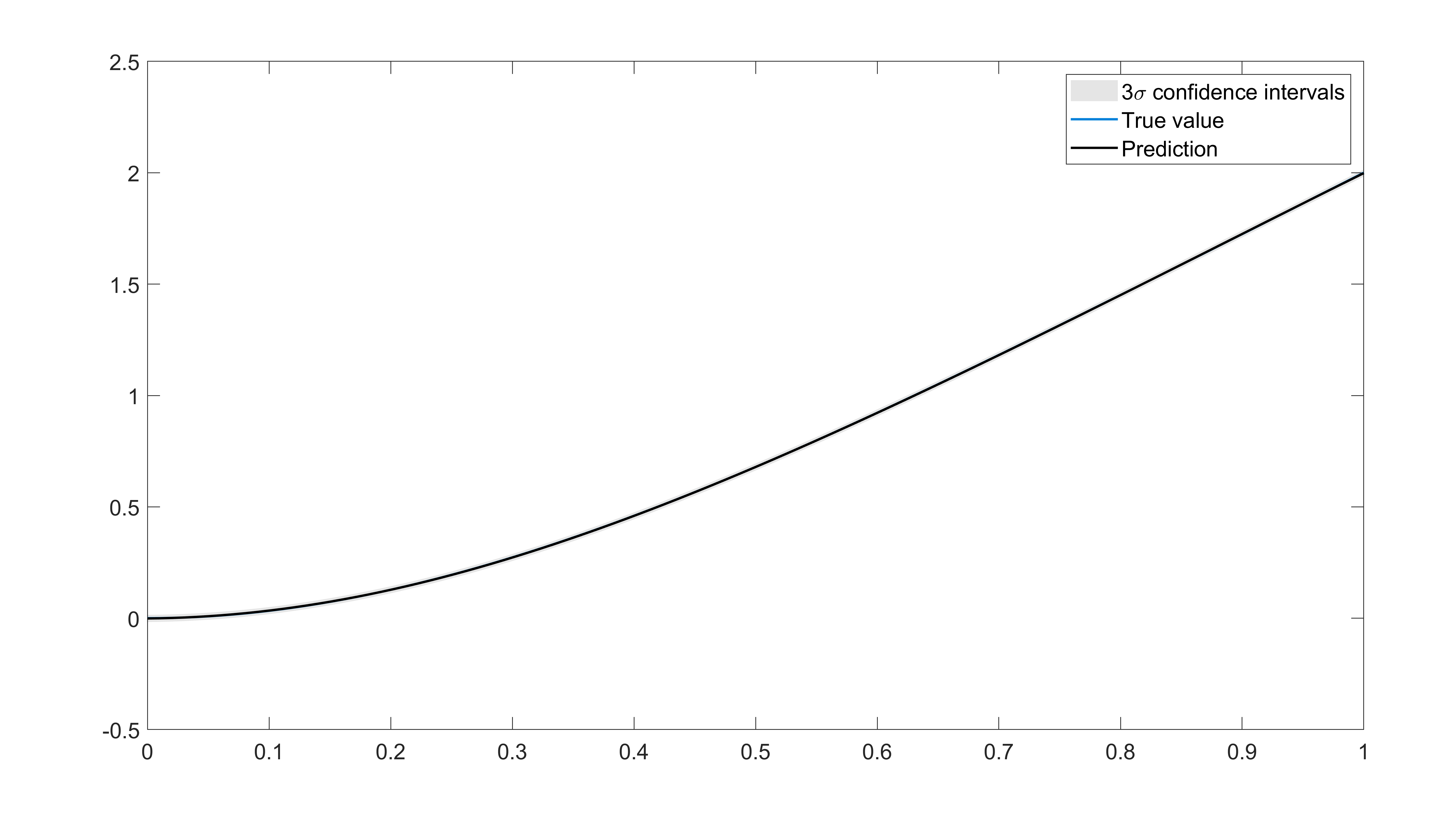}
		\caption{}
	\end{subfigure}
	\begin{subfigure}{0.45\linewidth}
		\includegraphics[width=\linewidth]{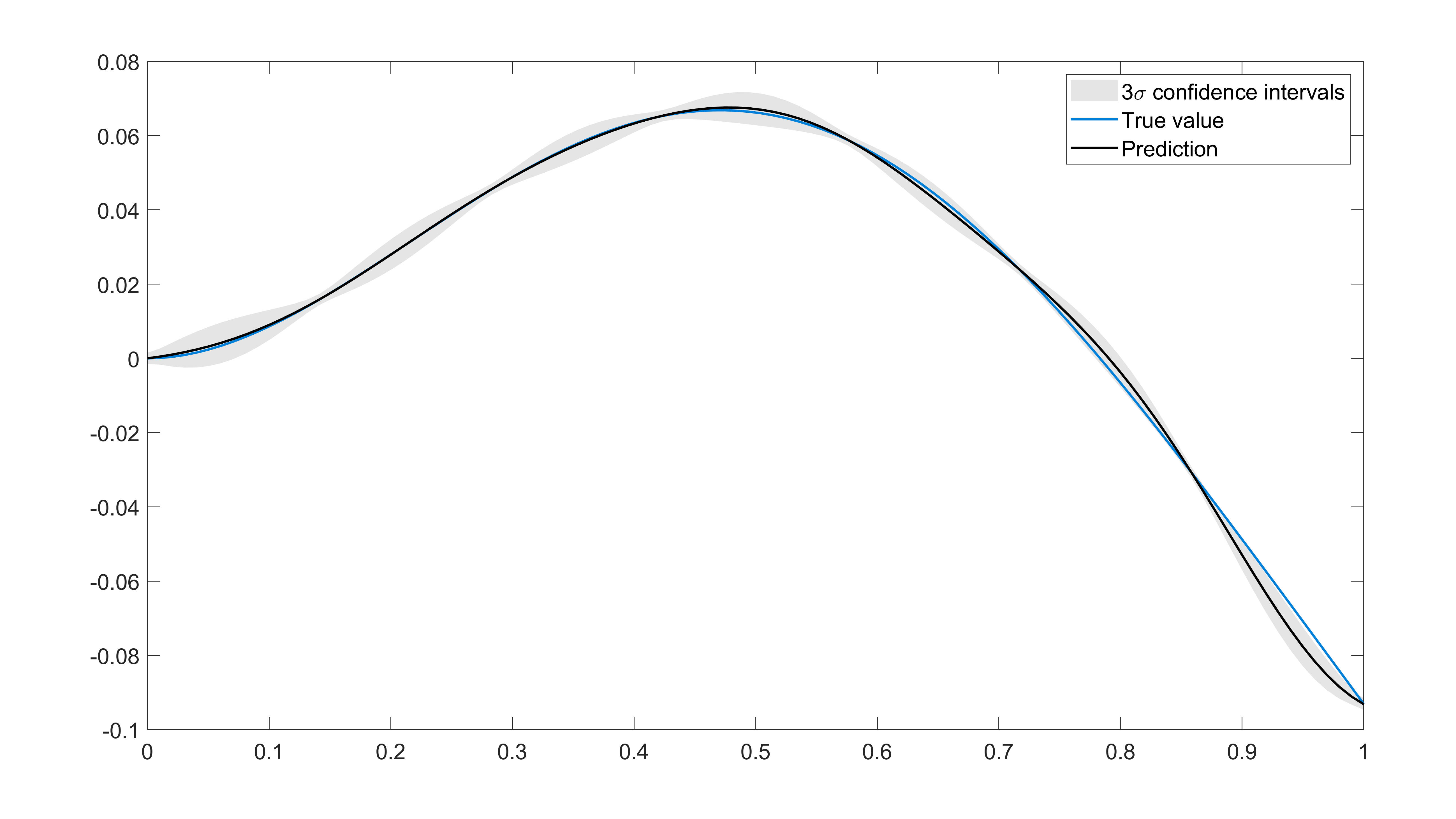}
		\caption{}
	\end{subfigure}
	\begin{subfigure}{0.45\linewidth}
	\includegraphics[width=\linewidth]{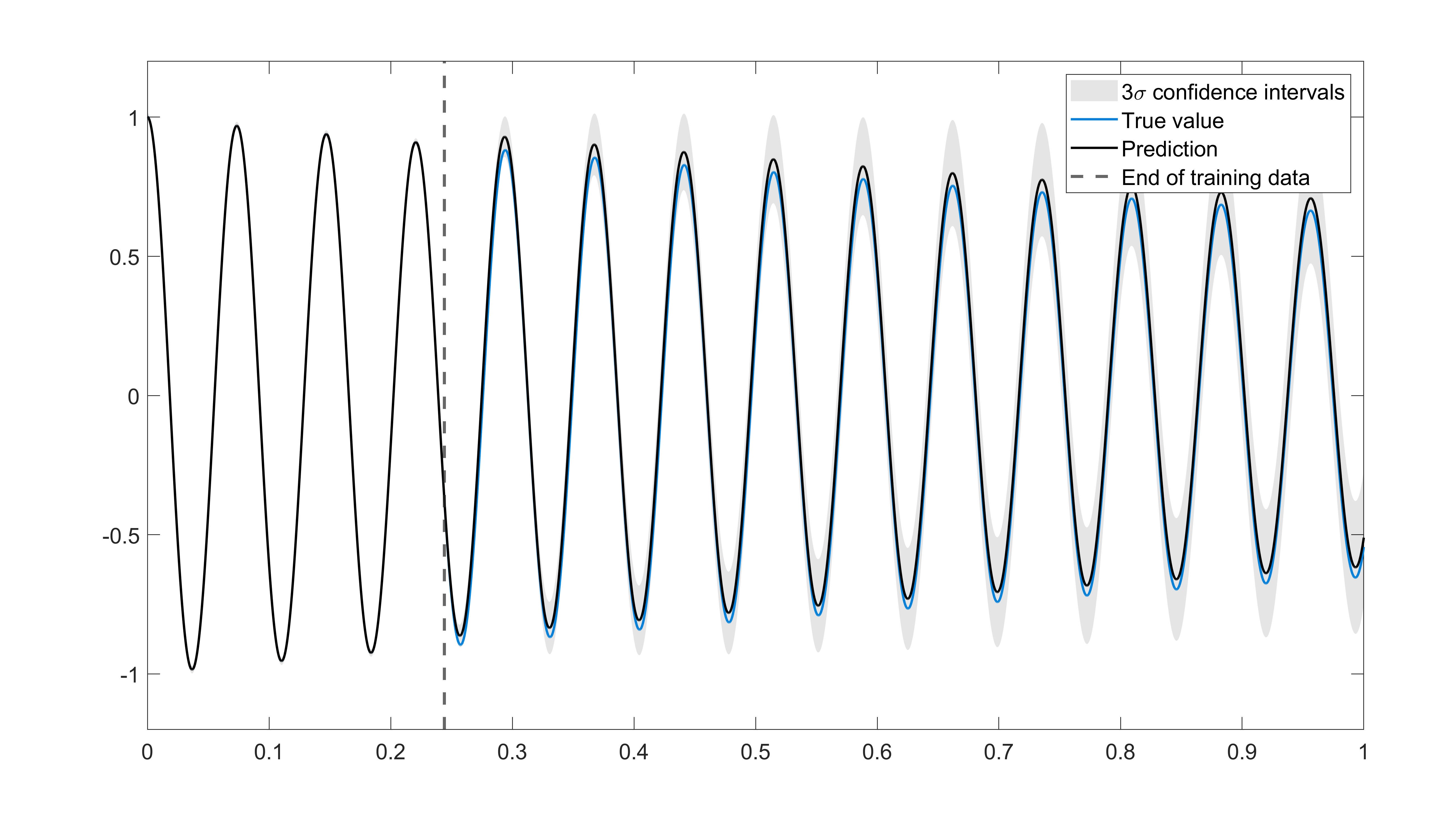}
	\caption{}
\end{subfigure}
\begin{subfigure}{0.45\linewidth}
	\includegraphics[width=\linewidth]{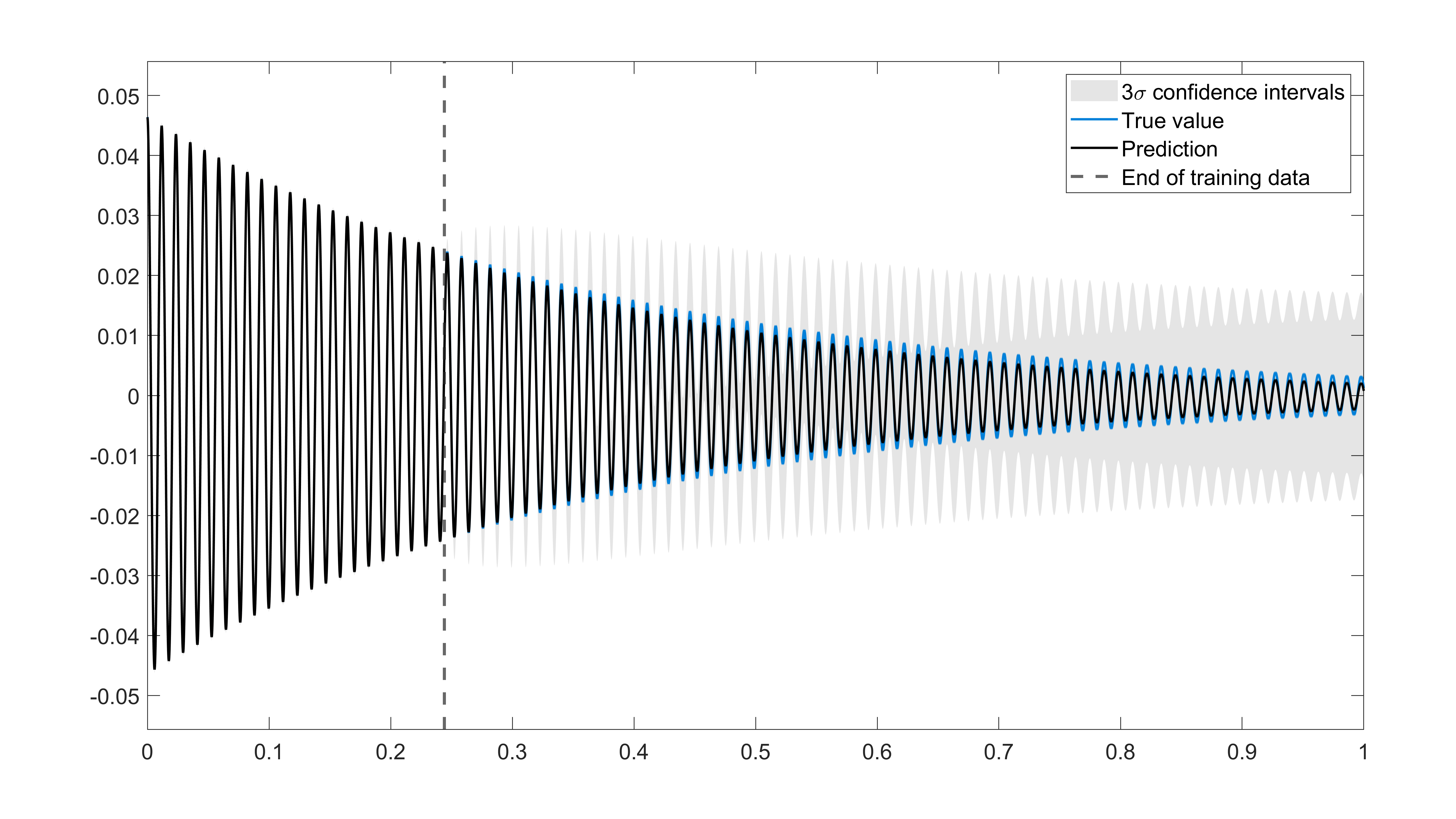}
	\caption{}
\end{subfigure}
	\caption{Hybrid covariance structure modelling the spatio-temporal behaviour of a vibrating beam - the spatial variation is assumed unknown. The predictions of the model are decomposed into the principal modes of the beam and shown here spatially (a,b) and temporally (c,d) for the first two modes. Normalised mean squared errors spatially are 0.002 and 0.225 (log loss -5.472,-3.088), with time domain errors 0.411 and 0.285 (log loss -3.520,-2.677) respectively \cite{pitchforth22incorporation}.}
	\label{fig:beams}
\end{figure}

Figure \ref{fig:beams} shows a simple illustration of this using a simulated cantilever beam under an impulse load - assuming here that we have no prior knowledge of the likely form of the spatial (modal) response. The GP is conditioned on a subsampled and truncated time history (\textit{1:2:T}) from eight points spatially distributed across the beam. The performance of the model in capturing the spatial temporal process is assessed by decomposing the beam response into its principal modes and comparing reconstruction errors across 100 spatial points and the full time history of the simulation (\textit{1:end}). Figure \ref{fig:beams} shows the GP prediction of the first two modes, where one can see that fidelity in the spatial and temporal domain are good. 

Knowledge of the system has been used in two ways here, firstly in developing the model structure (the kernel product) and secondly in accounting for the time domain behaviour through derived covariance. The relative importance of each is dependent on the availability of training data. Where fully sampled temporal and spatial data are available, a black-box counterpart should be comparable in performance so long as the selected covariance is sufficiently flexible/expressive. Knowledge of the separability of the domains has allowed good prediction here where limited data are available spatially, with the temporal knowledge becoming important where data are not fully sampled in time.


\subsection{Constrained covariance functions}
At the darkest end of the spectrum are problems and models where insights may be more general in nature, particularly where that knowledge cannot be used to derive generative equations. 

{\bf Example 4} $f({\bf x})\sim \mathcal{GP} (0,g(k_{ML}))$

This example looks at the problem of crack localisation in a complex structure using acoustic emission monitoring (acoustic emissions occur during the initialisation and growth of a cracks which may be detected and located through high frequency sensing \cite{jones2022bayesian,jones2023novel}). The task central to the localisation is to attempt to learn a map of how energy propagates through a structure from any possible crack location to a number of fixed sensors that are deployed to monitor acoustic emissions. The map is created by introducing a forced acoustic burst at each possible crack location (with a laser or pencil lead break), and then, in our case, producing a interpolative model that can be used inversely to infer location when a new emission is recorded. A large restriction limiting the use of such a localisation scheme is the need to collect training data across the structure, which can be time consuming and costly. Similarly, from a physical perspective, propagation of AE through a structure is complex (and costly) to model unless that structure is homogenous and of simple geometry. The proposed solution here is to follow a data-driven approach but with inbuilt information of the geometry/boundaries of the system, hopefully helping with both shortcomings.
 
\begin{figure}
	\centering
	\begin{subfigure}{0.45\linewidth}
		\includegraphics[width=\linewidth]{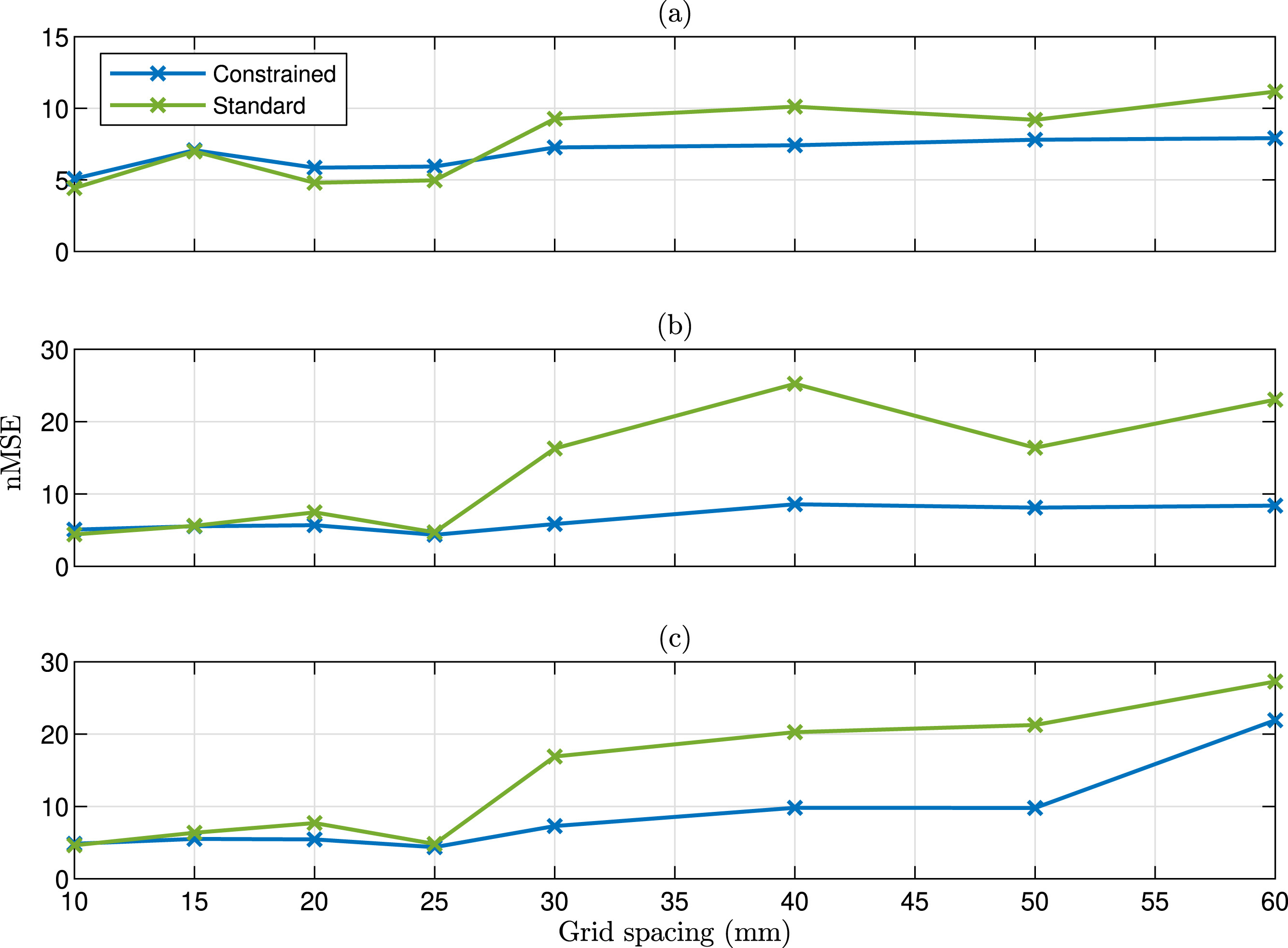}
		\caption{}
	\end{subfigure}
	\begin{subfigure}{0.45\linewidth}
		\includegraphics[width=\linewidth]{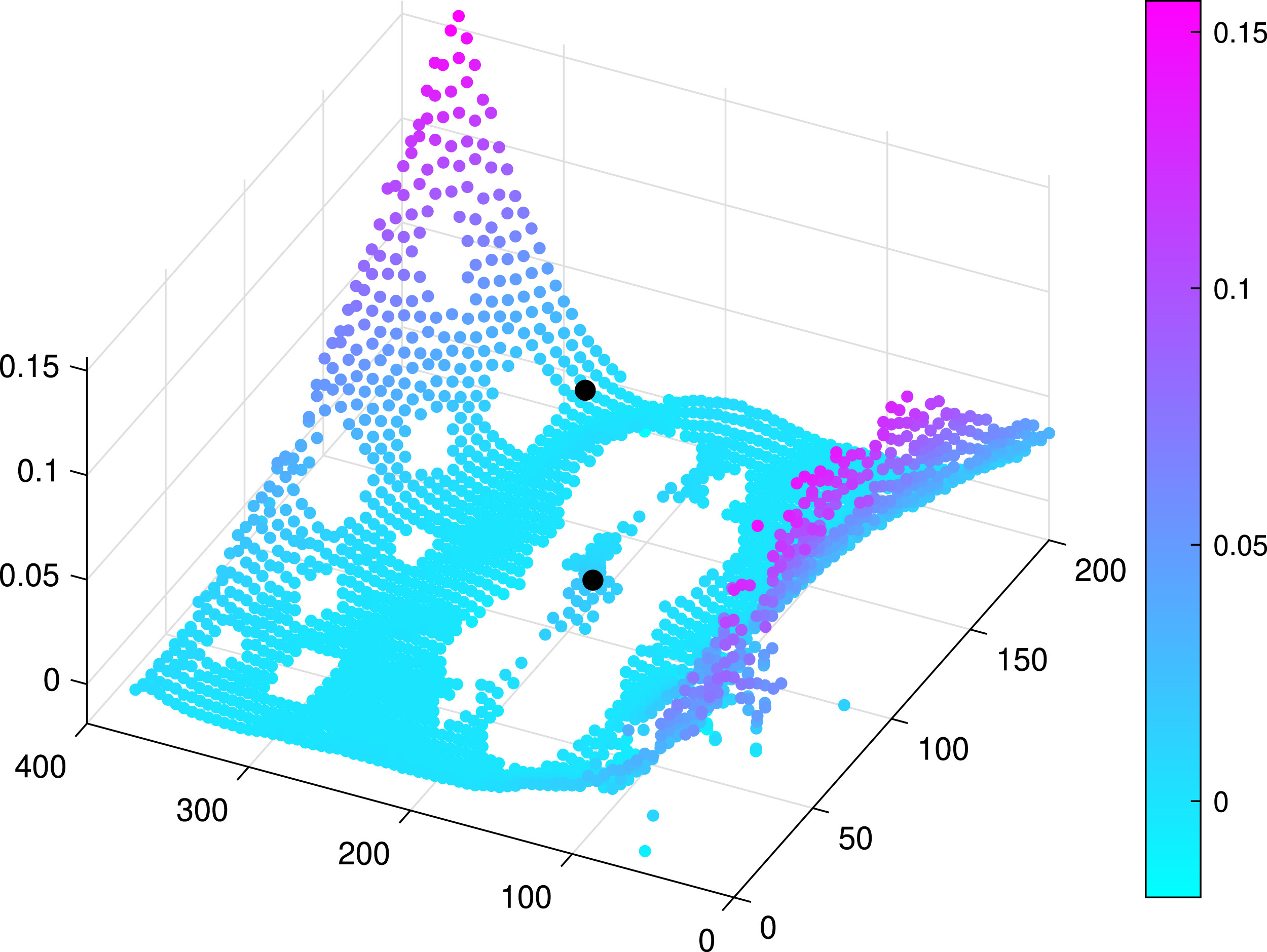}
		\caption{}
	\end{subfigure}
	\caption{Comparison of standard and constrained GP models for AE crack localisation \cite{jones2023constraining} where training data have been limited to the middle section of the plate. (a) compares model errors at decreasing training grid densities, with measurements at the boundaries included (top), partial boundary measurements (middle) and no boundary measurements (bottom). (b) shows an example of squared error difference between the two models across the plate, with the standard GP showing increased errors away from the training area.}
	\label{fig:cons}
\end{figure}

Figure \ref{fig:cons} shows how the use of a constraint to a standard machine learning covariance function can considerably lessen reliance on full data capture. Here a localisation map has been made of AE propagation through a plate with a number of holes in; a standard GP is compared to one where knowledge of the boundaries of the plate have been built into a sparse approximation of the covariance function \cite{jones2023constraining}. The two approaches perform similarly where training data are abundant, but we begin to see the gains of the introduced boundaries as training grid density decreases, and particularly so when parts of the plate are not mapped at all in the training phase.

\section{Discussion and conclusions}
\label{sec:discussion}



Each of the examples shown here has demonstrated how the introduction of physical insight into a GP regression has lessened reliance on conditioning data and has shown significantly increased model performance over black-box examples where training data are incomplete. 

The number of ways of accounting for knowledge within the GP framework provides parsimonious means to capture many of the different forms of prior knowledge that engineers possess and, in some cases, particularly when employing a mean function, for example, can be quite simply achieved. 

An additional benefit of many of the models shown is an enhanced interpretability, which is particularly important in the applied setting. In examples 1 and 3, the hyperparameters of the GP are the physical parameters of the system itself, opening the door to system identification in some cases.

Of the methods introduced/discussed, some are more familiar than others. The use of a mean function here takes the same approach as the bias correction community when a GP regression is used to account for discrepancy between a physical (often numerical) model and measurements of the real system \cite{gardner2021learning,kennedy2001bayesian}. Underlying the application shown here is the unspoken assumption that physics built into the mean function is trusted, thus allowing the consideration of extrapolation. Although the flexibility of the GP means that it is well able to account for potential bias, in its presence and in the case of incomplete data available for training, one is \textit{as} unable to place trust in the model across the operational envelope as one would be in the black-box case. In the case of incomplete data, this would suggest that only the simplest physics of which one is confident should be built in to the regression. 

The design of useful kernels is also naturally an area of interest for many, although often for different purposes to those explored here. Covariance design has been considered within the control community to improve the performance of machine learners for system identification tasks \cite{ljung2010perspectives,schoukens2019nonlinear,pillonetto2010new}. \cite{pillonetto2014kernel} provides a review in this context, where the focus is on the derivation of a covariance function that will act as an optimal regulariser for the learning of linear dynamical system parameters. Within the machine learning community, researchers attempting to develop more generic technology also look to physical systems to provide covariance functions useful for a broad class of problems \cite{wilson2013gaussian,parra2017spectral,tobar2015learning,van2017convolutional,higdon2002space,boyle2005dependent,alvarez2009latent,ross2021learning,mcdonald2021compositional}. These examples present very flexible models that are able to perform very well for a variety of different tasks. As discussed in the introduction, the motivation here is to build in system specific knowledge to lessen reliance on data capture. These models will only help with this tasks where the physics that inspired the general model is representative of the process of interest.

Including pertinent physical insight in a Gaussian process regression has most commonly been 
achieved via the multiple output framework, where relationships between 
multivariate targets are encoded in cross-covariance terms, including those studies already mentioned whilst discussing constraints \cite{solin2018modeling, wahlstrom2013modeling,jidling2018probabilistic}.  In \cite{cross2019grey} we adopt a 
multiple output GP to constrain a predictor using knowledge of physical 
boundary conditions for a structural health monitoring task. \cite{wan2019bayesian} shows a more comprehensive approach in this context, where the relationships between monitored variables are captured with the multiple output framework, different combinations of covariance functions are also considered. Notable contributions relevant here and also applied within an engineering context \cite{raissi2017machine,raissi2017inferring,raissi2018numerical}, use a differential operator to constrain multiple outputs to represent a system of differential equations. All of these works, which show significant improvement over an 
entirely black-box approach, adapt the standard machine learning covariance 
functions commonly used for regression. Expanding the possibility of directly derived priors in both mean and covariance as done here provides an opportunity (where available) to improve these models further.

The hybrid methods discussed in the medium (murky) grey section have been less well studied in this context and are the subject of ongoing work. The more complex interaction between physical and data-driven components has the potential to provide powerful models, although their use and interpretability will depend on architecture. When combining covariance functions over the same input domain, for example, the flexibility of the data-driven component will generally mean that hopes of identifiability are lost, whilst, nonetheless, still preserving predictive power. 

Finally, across disciplines, there are a growing number of examples now available demonstrating how knowledge of the boundaries or constraints of a system can be very helpful in the automated learning of their corresponding mapping \cite{coveney2020gaussian,swiler2020survey}, with many of the multiple output GP examples discussed above falling in this category. The flexibility of such models, alongside the opportunity to build in the simplest of intuitions will likely prove very popular in the future.

The main aim of this work was to provide a spectrum of potential routes for accounting for differing levels of physical insight within a regression context, using a Gaussian process framework. The examples employed here have demonstrated how the derivation path proposed can allow one to establish simple yet flexible models with components and/or (hyper)parameters linked to the physical system. This has been shown to be a desirable pursuit when models must be established without an abundance of training data - a common occurrence across engineering applications, where monitoring of our key infrastructure remains a difficult and expensive challenge.

\subsection*{Acknowledgements}
The authors would like to thank Keith Worden for his 
general support and feedback on this manuscript. Additionally, thanks is offered to 
James Hensman, Mark Eaton, Robin Mills, Gareth Pierce and Keith Worden for 
their work in acquiring the AE data set used here.  We would like to thank 
Ki-Young Koo and James Brownjohn in the Vibration Engineering Section at the 
University of Exeter for provision of the data from the Tamar Bridge. 

\subsection*{Funding statement}
The authors would like to acknowledge the support of the EPSRC, particularly 
through grant reference number EP/S001565/1, and Ramboll Energy for their 
support of SG and DP.

\subsection*{Competing interests statement}
The authors confirm that no competing interests exist.

\subsection*{Data availability statement}
Data availability is not applicable to this article as no new data were created or analysed in this study.

\subsection*{Author contributions - CRediT statement}
\textbf{EJC;} Conceptualisation (Lead), Methodology (Equal), Validation (Equal), Formal analysis (Equal), Investigation (Equal), Data Curation (Equal), Writing - Original Draft (Lead), Writing - Review \& Editing (Lead), Supervision (Lead), Funding acquisition (Lead)
\textbf{TJR;} Methodology (Equal), Validation (Supporting), Formal analysis (Supporting), Investigation (Supporting), Data Curation (Supporting), Writing - Review \& Editing (Supporting), Supervision (Supporting)
\textbf{DJP;} Methodology (Equal), Validation (Equal), Formal analysis (Equal), Investigation (Equal), Data Curation (Equal), Writing - Review \& Editing (Supporting)
\textbf{SJG;} Methodology (Equal), Validation (Supporting), Formal analysis (Supporting), Investigation (Supporting), Data Curation (Supporting), Writing - Review \& Editing (Supporting)
\textbf{MRJ;} Methodology (Equal), Validation (Equal), Formal analysis (Equal), Investigation (Equal), Data Curation (Equal), Writing - Review \& Editing (Supporting).

\bibliographystyle{unsrtnat}
\bibliography{grey_box_kernels}
\end{document}